\documentclass{article}
\usepackage[T1]{fontenc}

\usepackage{graphicx}
\usepackage{authblk}

\usepackage[natbibapa]{apacite}

\usepackage{flafter}
\usepackage{placeins}

\usepackage{longtable}
\usepackage{array}

\usepackage[hidelinks]{hyperref}
\usepackage[nameinlink]{cleveref}

\providecommand{\keywords}[1]{\textbf{\textit{Keywords:}} #1}

\title{RegCheck: A tool for structured comparisons between study registrations and papers}

\author[* 1,2]{Jamie Cummins}
\author[1]{Beth Clarke}
\author[1]{Ian Hussey}
\author[1]{Malte Elson}

\affil[1]{\small{Institute of Psychology, University of Bern}}
\affil[2]{\small{Bennett Institute of Applied Data Science, University of Oxford}}

\affil[*]{\small{Corresponding author: jamie.cummins@unibe.ch}}

\begin{document}
\maketitle

\begin{abstract}
Across the social and medical sciences, researchers recognise that specifying planned research activities in advance (i.e., ``registration'') has benefits for both the transparency and rigour of science. Despite this, evidence suggests that study registrations frequently go unexamined, limiting their utility. In a way this is no surprise: checking registrations against papers is labour- and time-intensive, requiring careful reading across formats and expertise across domains. Generative large language models (LLMs) are promising in enabling these checks, but ad hoc use of commercial chat models is unlikely to support standardised, reproducible, verifiable, and confidential processes. To overcome this, we present RegCheck (\url{https://regcheck.app}), an open-source, modular LLM-assisted tool for structured registration--paper comparison for researchers, reviewers, and editors from across the social and medical sciences. Importantly, RegCheck keeps human expertise and judgement in the loop by ensuring that (i) users are the ones who determine which features should be compared, and (ii) all judgements of (in)consistency between registrations and papers are supported directly by deterministic verbatim text segments which users can use to trivially verify all judgements and claims made by the LLM. We illustrate the utility of RegCheck with a worked example based on a hypothetical clinical trial, showing how the software can be used to support comparisons. We additionally discuss ongoing evaluation of RegCheck, as well as how to use the software fully locally using its command line interface to maximise privacy and security. By lowering the practical barrier to registration--paper comparison, RegCheck aims to make consistency checks a routine part of authors' workflows, peer review, and post-publication evaluation of research -- thereby maximising the potential and utility of study registrations.
\end{abstract}

\newpage
\section{Introduction}
The importance of registering study designs, materials, hypotheses, outcomes, and processing/analysis plans for the transparency and rigour of research has seen increasing recognition in the medical and social sciences in the last 20 years. In clinical trials, registration is a legal requirement according to the regulations governing many countries such as the United States \citep{noauthor_nih_nodate} and Europe \citep{noauthor_regulation_2014}. Even when not legally required, registration sees use across the sciences. In general medical research, the creation of registrations (or publication of study protocols) is growing in prevalence \citep{tan_prevalence_2019}; the same trend can also be seen in economics, with pre-analysis plans \citep{ofosu_pre-analysis_2023}, and in psychology, with preregistration (and more recently, Registered Reports, \citeauthor{chambers_past_2021}, 2021; \citeauthor{hardwicke_prevalence_2024}, 2024). Although different fields have different labels, norms, and literatures discussing these approaches, for the sake of consistency here, we refer to all of these using the term ``registration''.

In principle, these registrations improve the transparency of decision making and plans associated with research programs. This transparent record allows researchers to evaluate whether planned analytic choices were followed or not, thus facilitating the trustworthiness of scientific research more broadly \citep{hardwicke_reducing_2023}. Crucially, registrations enable the reader to distinguish which analytic decisions were made before and after observing the data, which can have implications for the veracity and severity of the reported statistical tests \citep{kaplan_likelihood_2015}. This provides researchers with critical information that they can use to evaluate the robustness of the results for themselves: in other words, researchers can more accurately assess the risk of bias and calibrate their confidence in the results accordingly \citep{hardwicke_reducing_2023}.

In practice, however, study registration is often more complex. Researchers may need to deviate from their registration for a variety of reasons \citep[see][for a discussion on when and how researchers might consider deviating from their registration]{lakens_when_2024}. Although some deviations could be unproblematic or trivial, others can seriously impact the severity of the test(s) of the predictions \citep{lakens_when_2024}. This is far from a distant concern---several meta-science studies across a variety of fields have found that deviations from registrations are common \citep{bakker_ensuring_2020, claesen_comparing_2021, goldacre_compare_2019, hahn_cross-sectional_2025, heirene_preregistration_2024, li_systematic_2018, ofosu_pre-analysis_2023, poole_systematic_2025, targ_meta-research_group__collaborators_estimating_2023, van_den_akker_selective_2023, van_den_akker_potential_2024}, and that these deviations are often not disclosed in the final paper \citep{van_den_akker_selective_2023, van_den_akker_potential_2024, targ_meta-research_group__collaborators_estimating_2023}\footnote{Another serious issue is that registrations are often underspecified and do not provide sufficient details to deviate from to begin with \citep{bakker_ensuring_2020, hahn_cross-sectional_2025, ofosu_pre-analysis_2023, poole_systematic_2025, van_den_akker_potential_2024}.}.

So how do these undisclosed deviations make it past peer review? The answer is probably quite simple. Checking registrations against the eventually published paper is a time-intensive and laborious process. Given peer reviewers are already overburdened \citep{adam_peer-review_2025}, the fairly substantial additional task of checking studies against their registrations seems unlikely to be done. Indeed, one study of articles published at \textit{PLOS} suggests that very few editors and reviewers engage with registrations during the peer review process \citep[see also \citeauthor{spitzer_registered_2023}, \citeyear{spitzer_stage_2023}]{syed_data_2025}. In a survey of reviewers for medical journals, only 34.3\% indicated they had examined the information on a trial registry within the past two years \citep{mathieu_use_2013}. This may also serve to explain why deviations are not reported by authors themselves: at least in some cases, it is plausible that authors are unaware of such deviations, given that checking for them is laborious.

While study registration offers a promising solution in principle to the risk of bias that can result from researcher degrees of freedom, in practice, the effectiveness of registration is seriously undermined by the work involved in conducting registration-paper comparisons. If registrations are never examined or evaluated, then we might be simply assuming that the risk of bias has been reduced when this may not be the case at all.

The recent advent of generative large language models (LLMs) offers the possibility of automating portions of the work involved in registration-paper comparisons, assisting reviewers in a task that they seem to avoid (note that we refer to ``registration-paper'' or ``registration-publication'' comparisons throughout this manuscript, but of course such processes can be done with drafts of manuscripts prior to publication or submission to a journal). Indeed, the simplest form of this automation is already accessible to the average researcher: upload both documents to a commercial LLM and prompt it to compare them. As we argue in the next section, however, this ad hoc approach cannot support robust research evaluation. 

What is needed instead is an approach which ensures consistency and standardisation in prompt content, comparison logic, and output across calls and users, while also ensuring that expert knowledge is applied by (i) allowing researchers to specify which dimensions should be examined, and (ii) facilitating researchers' assessment of discrepancies. To meet these needs, we have developed \href{https://regcheck.app}{RegCheck}: a tool which automatically extracts the most relevant text from registrations and papers, easily facilitating comparison between these sources. It provides a pipeline that parses registration and publication texts, lets users specify what dimensions they want to check, and produces structured, auditable reports which are shareable via unique links. By lowering the difficulty barrier for checking registrations without removing expert oversight, RegCheck aims to make consistency checks a routine part of scientific workflows.

RegCheck is a single instance of a broader class of emerging meta-scientific tools: those whose purpose is to provide (semi-)automated comparisons between academic research objects for which consistency is expected and deviations are informative. We refer to the abstract architecture underlying this class as \emph{Parity}. The remainder of this paper first considers why ad hoc use of LLMs cannot support robust registration-paper comparison. We then describe Parity and the IDEA framework that organises it, and introduce RegCheck as a concrete implementation of Parity for registration-paper comparison, including a worked example of RegCheck in action. Finally, we turn to how RegCheck is used in practice, how we are evaluating it, and the research agenda behind it.

\section{Why Not Just Ask an LLM?}
\label{sec:why-not-llm}
Before describing the workflow itself, it is worth engaging with a simple question: since generative LLMs are already widely accessible, why is a dedicated tool needed at all? Any researcher can already upload a registration and a paper to a commercial chat model and prompt it to compare them; in one sense, the automation problem is ``solved''. However, such an ad hoc approach cannot support comparisons that are standardised, reproducible, verifiable, and confidential---all properties that formal research evaluation demands. Below we discuss six problems associated with simply using a standard LLM chat interface.

\begin{enumerate}
\item \textbf{Analytic flexibility.} Even for an identical underlying task, LLM output can vary substantially with the phrasing of the prompt, the formatting of the documents, and the model's input parameters. For instance, the difference between the presence or absence of a colon in a required output format can shift LLMs' accuracy in a categorisation task by up to 76 percentage points \citep{sclar_quantifying_2023}. Additionally, other analytic decisions (e.g., choice of values for temperature/reasoning effort, the granularity of prompt requests, system prompt content, and a range of other factors) can substantially influence the output of LLMs \citep{cummins_threat_2025}. Without a standardised workflow, there will almost certainly be a substantial degree of variation between users in terms of which model they use, how they prompt it, memories or content that are present in their conversation history, and a range of other influential factors. All of these factors mean that any two researchers prompting an LLM may very well receive substantially different outputs. This also serves as a barrier to formal evaluation of the performance of these tools: it is not possible to estimate how accurate an LLM \emph{in general} is at comparing preregistrations and papers, because this performance is necessarily tied to this myriad of factors and analytic decisions that a researcher can make (or that are  obscured when using commercial chat interfaces, see point 2). Robust usage and evaluation therefore requires standardised inputs, including fixed prompt structures and definitions, consistent procedures for supplying source material to the model, and easily definable hyperparameter values.
\item \textbf{Commercial chat interfaces are not appropriate for research evaluation.} In practice, ad hoc use almost always means that consumer chat products are used. Put simply: these interfaces are not suitable for use as a formal, standardised research tool. This is the case firstly because many of the influential analytic decisions mentioned in the previous point are typically gated off from the user when using commercial chat interfaces. In other words: chat interfaces typically do not allow users to modify choices related to temperature/reasoning effort, full system prompt content, and other influential factors \citep{higgins_recommendations_2026}. Worse still, these settings are not merely fixed by the LLM provider, but unstable: for example, versions of the deployed model, system prompts, and server-side processing steps may be changed without notice to, or control by, users. 
\item \textbf{Data privacy.} Registration-paper comparison routinely involves unpublished manuscripts, including materials handled under the confidentiality expectations of peer review. When files are uploaded to a commercial chat interface, providers typically reserve the right, under default settings, to use this data for model training; usage agreements for data submitted via APIs are generally more restrictive. Moreover, commercial chat interfaces provide access only to their provider's own models: open-weight models (which can be hosted by parties with no commercial stake in the data, by institutions, or on fully local infrastructure) are largely inaccessible through them. A workflow built around chat interfaces therefore forecloses the strongest available privacy measures, such as running an identical comparison entirely on local infrastructure.
\item \textbf{Unwanted dependency among comparisons.} Because of the nature of the next-token prediction architecture that constitutes large language models, subsequent predictions for outputs (in a chat context) are influenced by previous predictions. For example, suppose a researcher prompts an LLM in the chat interface to compare a registration with a paper in terms of hypotheses and statistical analyses. The nature of next-token prediction means that the response for ``statistical analyses'' will necessarily be influenced by the output for the previously-answered ``hypotheses'' dimension. Asking a model to make comparisons on multiple such dimensions of interest in a single pass necessarily introduces dependencies among the resulting judgements because output is generated sequentially - for example, if a model states that the registration and paper do not match based on hypotheses (when in reality they do), then this may also increase the likelihood that the model states that the registration and paper do not match based on the statistical analyses, given that hypotheses and statistical analyses may often be related. Errors can therefore propagate across dimensions, and dimension-level accuracy can no longer be evaluated independently. This makes both reproducibility and meaningful evaluation substantially harder to achieve.
\item \textbf{Conflation of distinct tasks.} Simply asking an LLM to make these comparisons delegates a number of distinct tasks to the LLM (i.e., ingestion, definition of dimensions of interest, extraction, and adjudication; see next section). However, this comes at the cost that other tools (including human input) may be more appropriate for some of these steps than others. It also means that it is difficult to evaluate the independent performance of the LLM on each of these component steps, because they are necessarily tied to one another (in a manner similar to point 4 above). 
\item \textbf{Verifiability.} Large language models will, by and large, provide a response to any query that does not breach their safety guidelines---including in cases where the response is incorrect (``hallucinations''). Any LLM-based research tool therefore needs to make the model's claims trivially verifiable. Although some LLM products provide citations for some claims, this is unsystematic and not guaranteed to occur in every case. If researchers must go through the documents themselves to identify where the LLM's claims come from, this is no more efficient than performing the comparison manually in the first place.
\end{enumerate}

Importantly, none of the above-mentioned problems serve as arguments against using LLMs for registration-paper comparison in general. Rather, these arguments highlight the need for the use of LLMs in a standardised, transparent, and modularised workflow which clearly delineates the different steps required for the comparison, as well as which tool should serve to perform each of those steps. Breaking down such workflows in this manner is the goal of the IDEA framework and its software implementation \textit{Parity}, which we now discuss.

\section{Parity and the IDEA Framework}
Parity is the more general software workflow of which RegCheck is an instance. It is not itself a single end-user tool, but rather a generic architecture that can be appropriated to implement different (meta-)scientific consistency-checks in a way that keeps inputs, evidence retrieval, and judgement steps conceptually distinct. The motivation for this abstraction is that many scientific evaluation tasks involve checking consistency between research artefacts. Registration-paper comparison is one example, implemented in RegCheck, but the Parity workflow can be used to assess the consistency between other pairs of research artefacts such as comparing manuscripts and external reporting standards (e.g., ARRIVE, PRISMA), or comparing manuscripts' natural language descriptions of their analyses with the code implementing the analyses \citep{cummins_jamiecumminscodebot_2026}. 

Parity is organised around a four-step conceptual framework that structures how to think about, break down, and approach (meta-)scientific comparison, extraction, and adjudication questions between research artefacts. We refer to this as the IDEA framework:
\begin{enumerate}
\item \textbf{Ingestion.} Collect all documents relevant to the specific task at hand, and parse them in ways that maximise their utility and clarity for that particular task. In practice this may involve interfacing with APIs and handling JSON responses (e.g., on ClinicalTrials.gov), XML extraction, PDF parsing, document-layout models such as GROBID, reference stripping, section filtering, or other deterministic preprocessing needed to transform scientific materials into task-ready text.
\item     \textbf{Definition.} Specify along which dimensions the scientific materials should be evaluated, compared, or checked. This step is explicitly human-defined: users (who may be reviewers, editors, or the authors themselves) define the dimensions of interest, whether these are reporting-guideline items, preregistered sample-size commitments, outcomes, or any other target of interest to the user.
\item \textbf{Extraction.} Represent documents and dimension definitions in a form that supports targeted retrieval, and then extract the most relevant evidence for each dimension. In practice, this is most frequently done by the use of semantic embeddings models, which represent semantic information through numeric vectors. Text extraction is then done by splitting scientific documents into chunks, embedding these chunks and dimension definitions specified in the previous step, and querying the similarity of each chunk with the dimension of interest.
\item \textbf{Adjudication.} Make a judgement for the task at hand on the basis of the extracted evidence. This is the stage at which generative LLMs are most naturally used, because the system must synthesise multiple excerpts and render a task-specific judgement, such as whether a registration and paper are consistent on a given dimension.
\end{enumerate}

In addition to providing a scaffolding for researchers to think about how to decompose their meta-scientific workflow into distinct steps, the IDEA framework also helpfully clarifies that different steps in the framework involve different tools or sources of input. This also modularises the backend code, increasing its reusability in other open-source scientific projects and allowing individual elements of the Parity workflow to be replaced as better solutions emerge (e.g., for the pernicious problem of text extraction from PDFs, with which many scientific software projects contend). This separation of the four steps, both conceptually and in code, also enables more fine-grained validation of any given Parity implementation: the validation of LLM-based Adjudication is distinct from the generic difficulties of Ingestion (such as PDF text extraction), and validation metrics can be designed to identify which steps succeed or fail.

\begin{figure}
    \centering
    \includegraphics[width=0.5\linewidth]{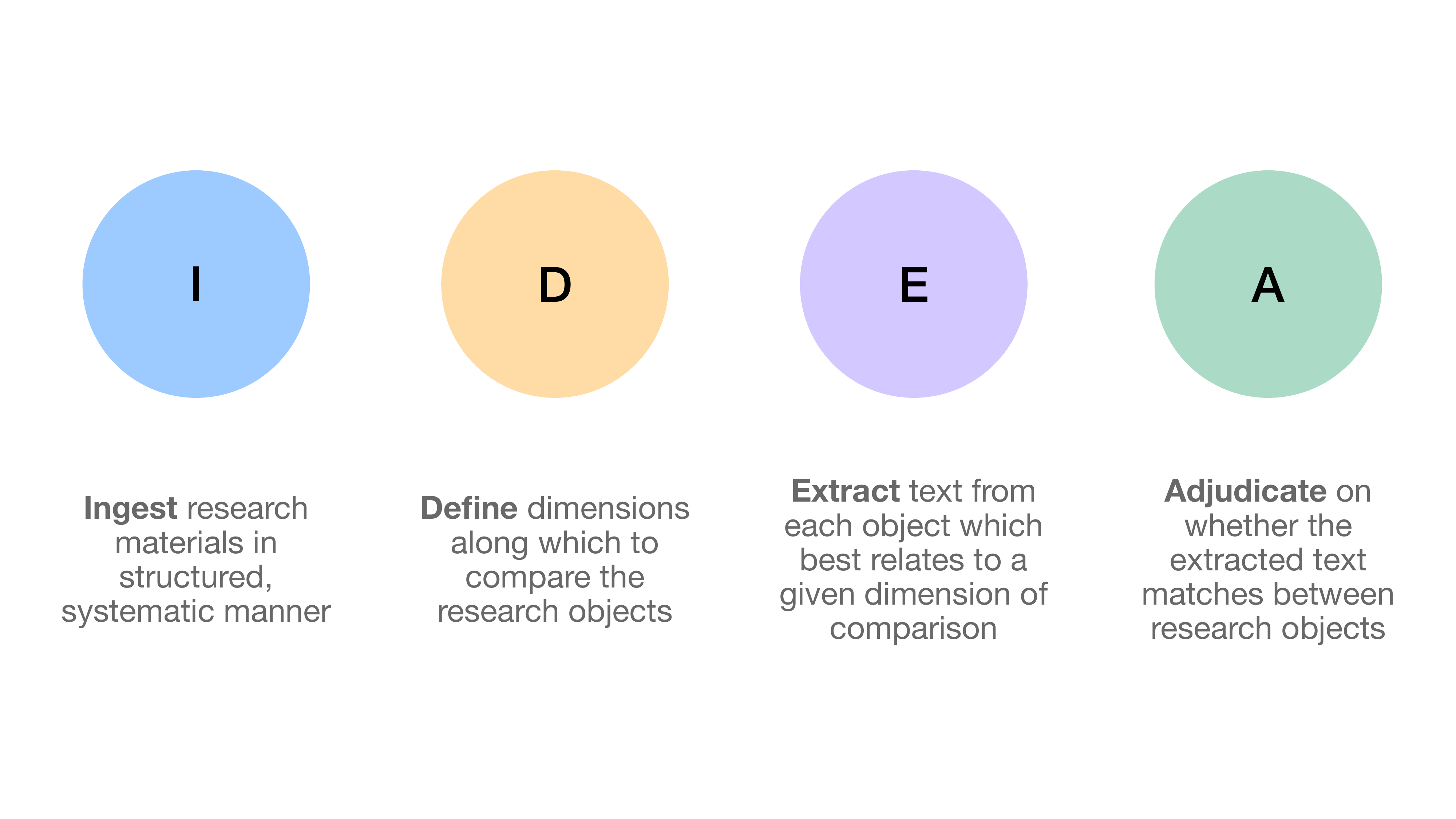}
    \caption{An illustration of the abstract sequence and breakdown of the IDEA framework.}
    \label{fig:idea-framework}
\end{figure}

Most critically, the IDEA framework disambiguates where generative LLMs are and are not used in these processes. Within the canonical IDEA workflow: Ingestion is primarily a matter of conventional approaches to document parsing; Dimension definition involves entirely human input; and Evidence extraction is based on embeddings models. Generative LLMs are by default only invoked at the Adjudication step, where synthesised judgement is actually required.  With the scaffolding of Parity and the IDEA framework in mind, this should serve to clarify to readers that RegCheck is not merely ``an LLM that reads and compares two documents'', but is instead a modular workflow that takes a systematic approach to the task of registration-paper comparison (see \Cref{fig:idea-framework}).

This separation of roles makes the workflow easier to inspect, evaluate, optimise, and modify, and helps to localise sources of error when they arise. Additionally, for users who are sceptical or reticent to use generative LLMs, this does not render the entirety of Parity workflows inaccessible, since the LLM is only used at the final stage of adjudication, and the preceding stages can be used without it (e.g., one could use the evidence extraction stage to retrieve relevant text excerpts for each dimension, and then make their own human judgement on the basis of that evidence without using an LLM at all).

RegCheck is therefore best understood as an implementation of Parity which is developed specifically for registration-paper comparison. With this context in mind, the main focus of this paper is to introduce and describe RegCheck itself, in particular its use case, workflow, report structure, and potential value for authors, reviewers, editors, and meta-scientists.

\section{The RegCheck Workflow}
\begin{figure}
    \centering
    \includegraphics[width=0.5\linewidth]{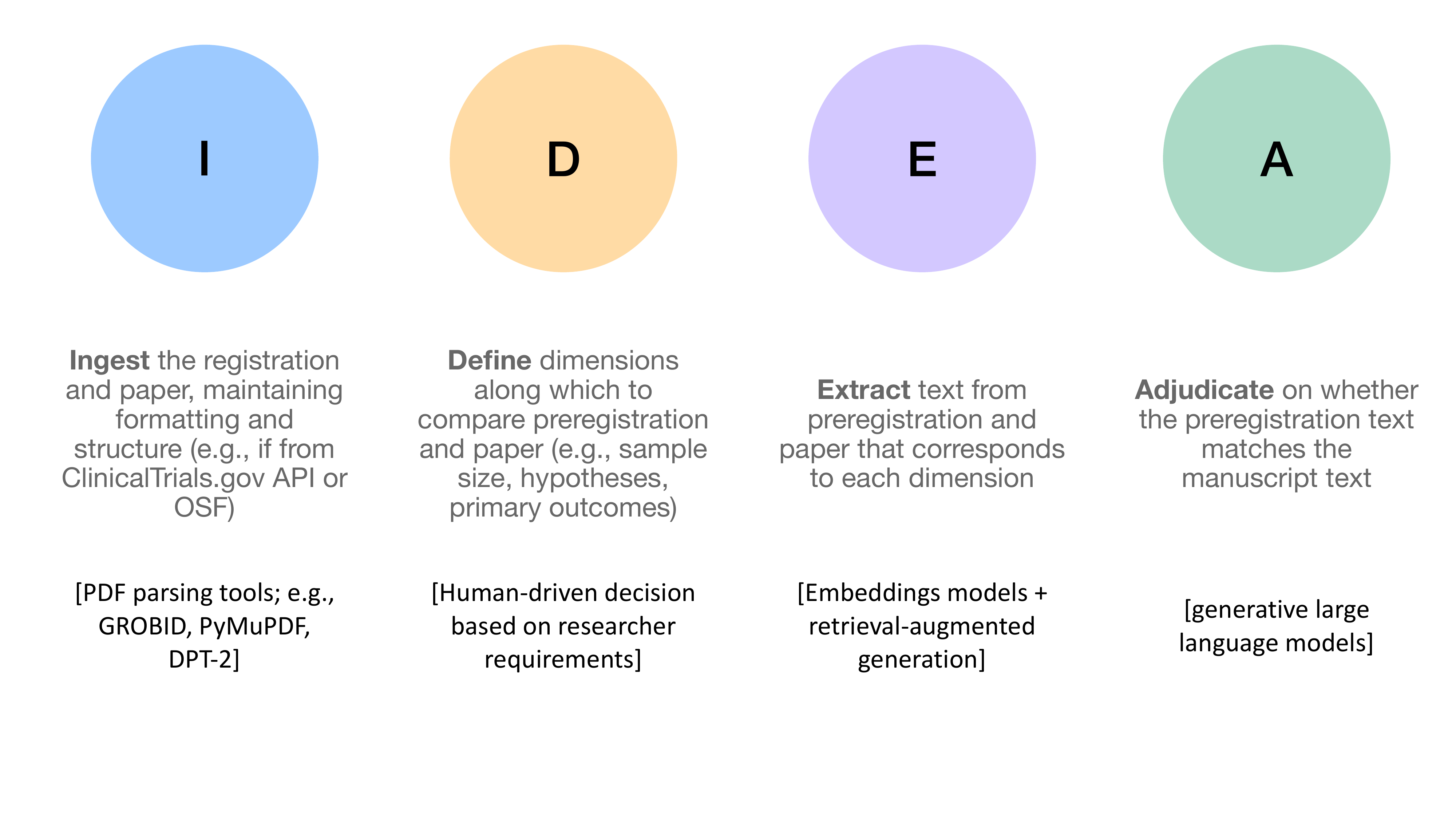}
    \caption{An illustration of the application of the IDEA framework specific to RegCheck's implementation, including clear delineation of the tools required for each step of the workflow.}
    \label{fig:regcheck-idea-framework}
\end{figure}

RegCheck operationalises the four IDEA steps through a six-stage backend pipeline. These additional stages reflect specific modifications to the framework that serve to better facilitate the specific task of registration-paper comparison. Stages 1-2 correspond primarily to \textbf{Ingestion}, Stage 4 to \textbf{Definition}, Stage 3 together with the retrieval component of Stage 5 correspond to \textbf{Extraction}, and the judgement component of Stage 5 corresponds to \textbf{Adjudication}. Stage 6 is an additional layer focused on reporting, which is not a core part of the IDEA framework but is important to ensure the interpretability and ease of use of the software (see \Cref{fig:regcheck-idea-framework}).

\begin{enumerate}
\item \textbf{Ingestion.} The user specifies a registration and paper. Registrations can be provided in a number of ways: for example, by directly uploading a registration file (e.g., PDF), referencing an entry in a public registry (e.g., ClinicalTrials.gov), or by linking to a hosted registration (specifically the Open Science Framework, provided the link resolves to either a registration's structured responses or an attached file). Papers are accepted in a range of document formats. Where a paper is supplied as a PDF, it can be parsed using specialised document-layout models (such as GROBID; \citep{grobid_grobid_2025}, or DPT-2; \citep{huang_document_2025}), with an automatic fallback to alternative extractors when the selected primary parser cannot recover usable text (e.g., from a scanned or atypically typeset document). Registrations obtained from registries or hosted repositories are handled through their structured-data interfaces rather than PDF parsing. Critically, all of these parsing approaches are engineered to ensure that relevant text content maintains its integrity and can be properly handled in the subsequent steps.
\item \textbf{Preparation.} Reference sections are stripped and layout is normalised. If the paper is indicated to consist of multiple studies, then the paper is further processed with irrelevant studies removed, and only the general introduction, the relevant study, and general discussion are retained. When the focal study refers back to details described elsewhere in the paper (e.g., ``Our procedure was the same as the previous experiment''), RegCheck applies additional follow-up extraction logic so that those antecedent details can be surfaced for comparison.
\item \textbf{Embedding and indexing.} All content from both documents is split into chunks using a token-based, sentence-aware method. Chunks are kept relatively short in length so that retrieved excerpts are tight, quotable, and matched to the types of extractions that humans would typically provide. Chunk boundaries are aligned to both sentence endings and section headings, so that (for example) a methods detail is not embedded together with adjacent results text. Consecutive chunks overlap by a modest amount, realised as whole trailing sentences, so that important information straddling a boundary still appears intact in at least one chunk. The exact chunk size and overlap are configurable parameters via the RegCheck API and CLI. 
Embeddings are then computed for these chunks using a general-purpose semantic embedding model to support retrieval-augmented extraction. Because the embedding step is provider-neutral, the embedding model can be swapped -- including for a locally or institution-hosted model -- so that registration and paper text can be embedded without leaving a trusted environment (indeed, local embeddings model usage is supported for local use within the CLI). These vector representations form the basis of the evidence-extraction process.
\item \textbf{Definition.} Users either select a built-in set of pre-defined dimensions, or define their own by providing a label and a definition for each dimension. The pre-defined dimensions are organised into discipline-specific presets (for example, distinct dimension sets for psychology, clinical/medical, economics, and preclinical/animal research). Each of these sets contains default definitions that are specific to that field's reporting conventions (for instance, clinical/medical defaults have a specific focus on primary/secondary outcomes), and this set of presets will be further developed over time. Definitions are used both for text extraction and for better-specified adjudication prompts.
\item \textbf{Extraction and adjudication.} For each dimension, RegCheck:
\begin{itemize}
    \item Retrieves the most relevant evidence excerpts from each source using dense embedding retrieval. It forms a single query from the dimension label plus its definition (user-provided or a built-in default), embeds this query, and scores every registration and paper chunk by cosine similarity. The number of excerpts retained scales with the length of each source (so that a short, single-study text is not over-pruned) and is complemented by a hybrid keyword search that promotes high-value chunks which a dimension's keywords match but which ranked just outside the similarity cut (e.g., a sentence which explicitly mentions a keyword like ``sample size'' or ``exclusion''). This process is the means by which RegCheck provides users with direct quotes from the registration/paper for their own inspection. RegCheck also uses these extracted quotes when constructing the adjudication prompt, while additionally expanding each retrieved chunk into its neighbouring chunks (a ``small-to-big'' approach) so that the model is also provided with information about those chunks in-context. 
    \item Sends the retrieved excerpts to a user-selected LLM, which produces concise summaries of the dimension-relevant registration and paper content (referencing the excerpt IDs).
    \item Produces a deviation judgement from the LLM using the same evidence, encoding whether a deviation exists, the materials are consistent, or the evidence is insufficient to judge, together with a brief supporting explanation. These outcomes are surfaced to the user as decisions of \texttt{Deviation},  \texttt{No deviation}, or \texttt{Insufficient evidence}. By default, each comparison dimension is adjudicated in its own call, so that the output for one dimension does not contaminate another (see the discussion of \textit{Append outputs} below).
\end{itemize}
\item \textbf{Reporting.} RegCheck renders an interactive, shareable report associated with a unique RegCheck report ID. The report comes with two views: Overview, and Evidence. \textit{Overview} presents, for each dimension, a colour-coded consistency judgement (Deviation, No Deviation, Insufficient Evidence) together with a rationale for this judgement and summaries of the content from the study registration and paper. Summaries and deviation judgement texts also directly cite specific 
excerpts from each document, allowing users to easily verify the claims made. \textit{Evidence}, on the other hand, offers a much more detailed view, showing both of the source documents (i.e., registration and paper) side-by-side, with relevant sections from a given dimension highlighted. The purpose of this view is to provide users with an easy way of directly identifying the content most relevant to that specific dimension within each document. In this way, users can more easily conduct manual comparisons between the two documents, as well as verify the specific outcome comparison provided by RegCheck. RegCheck reports can be exported (e.g., as HTML files), as well as shared either publicly or privately (as determined by the user who generated the report). \Cref{fig:overview-report} and \Cref{fig:evidence-report} provide illustrations of the appearance of a RegCheck report.
\end{enumerate}

\begin{figure}
    \centering
    \hspace*{-\oddsidemargin}\hspace*{-0.75in}
    \includegraphics[width=0.95\paperwidth]{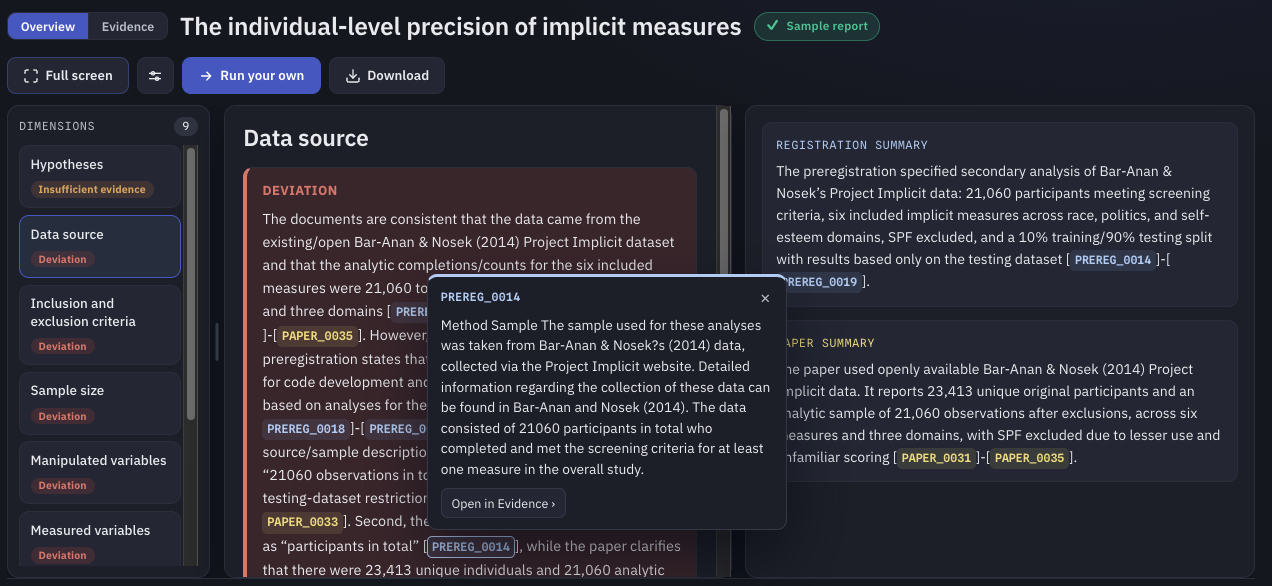}
    \caption{A screenshot from the Overview section of a RegCheck report. The left panel specifies the dimension of interest, along with the associated judgement for this dimension; the central panel provides the rationale for this deviation judgement; the right panel provides overviews of the findings from the registration and paper. In this screenshot, one of the quotes specified in the report has been clicked, revealing the verbatim content of the quote to the user.}
    \label{fig:overview-report}
\end{figure}
\begin{figure}
    \centering
    \hspace*{-\oddsidemargin}\hspace*{-0.75in}
    \includegraphics[width=0.95\paperwidth]{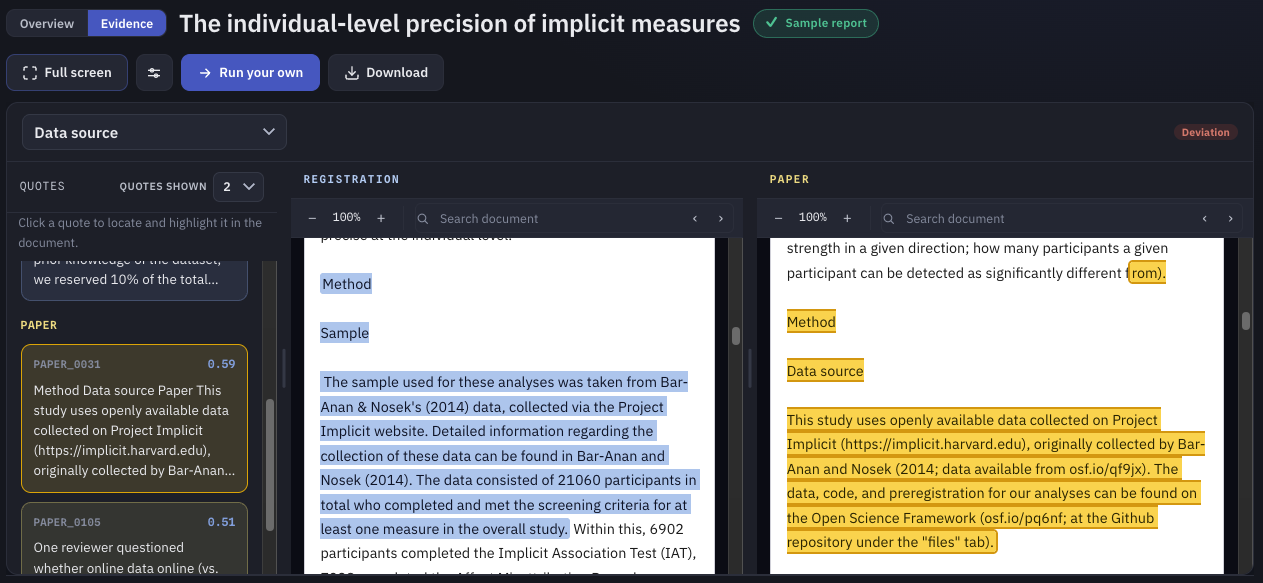}
    \caption{A screenshot from the Evidence section of a RegCheck report. The left panel displays the quotes most relevant to the specific dimension under examination from both the registration and paper. The central and right panels provide renders of the text from the registration and paper, respectively. When a quote of interest in the left panel is clicked, the report automatically navigates in the registration or paper to the location of the quote of interest.}
    \label{fig:evidence-report}
\end{figure}

\FloatBarrier
\section{A Worked Example}
\label{sec:worked-example}
To make the workflow concrete, we walk through an illustrative example showing how RegCheck can be used. The content of the trial registry/paper below is hypothetical, but serves to show how RegCheck can provide a useful and verifiable report that can facilitate registration-paper comparison.

Consider a randomised, placebo-controlled trial of a once-weekly drug, added to standard care, for adults with type 2 diabetes. A reviewer has the manuscript and the trial's public registry entry, and wants to check them against one another using the RegCheck web application. Using the clinical-trial flow, the reviewer enters the trial's ClinicalTrials.gov identifier and uploads the manuscript, then opts for the clinical/medical dimension preset and runs the comparison. RegCheck retrieves and structures the registration directly from the registry, parses the manuscript, and for each dimension, chunks and embeds both sources (i.e., the registration and manuscript), retrieves the most relevant excerpts from each, summarises them, and adjudicates whether the paper deviates from what was registered. \Cref{tab:worked-example} shows a simplified version of the resulting hypothetical report for all eleven dimensions of the clinical/medical preset.

\begin{table}[!htbp]
\footnotesize
\renewcommand{\arraystretch}{1.5}
\setlength{\tabcolsep}{4pt}
\caption{An illustrative RegCheck report for the hypothetical type 2 diabetes trial, showing all eleven dimensions of the clinical/medical preset. Registration excerpts are drawn from structured registry fields; evidence identifiers (e.g., \texttt{PREREG\_0011}) are the handles a reader clicks to jump to the highlighted excerpt in the source. All quotations are constructed for illustration and are not derived from existing manuscripts or registrations.}
\label{tab:worked-example}
\hspace*{-\oddsidemargin}\hspace*{-0.10in}
\begin{tabular}{p{0.25\linewidth} p{0.37\linewidth} p{0.37\linewidth} p{0.37\linewidth}}
\hline
\textbf{Dimension} & \textbf{Registration (excerpt)} & \textbf{Paper (excerpt)} & \textbf{RegCheck verdict} \\
\hline
\textbf{Eligibility -- inclusion criteria} &
``Adults aged 18--75 years with type 2 diabetes and a baseline HbA1c of 7.0\%--10.0\%.'' [\texttt{PREREG\_0004}] &
``Eligible participants were adults aged 18--75 years with type 2 diabetes and a screening HbA1c of 7.0\%--10.0\%.'' [\texttt{PAPER\_0009}] &
\textbf{No deviation.} The reported eligibility matches the registered criteria. \\
\hline
Eligibility -- exclusion criteria &
``Exclusion: estimated glomerular filtration rate below 30 mL/min.'' [\texttt{PREREG\_0006}] &
``Patients with an estimated glomerular filtration rate below 45 mL/min were excluded.'' [\texttt{PAPER\_0011}] &
\textbf{Deviation.} The renal-function exclusion threshold was tightened from 30 to 45 mL/min. \\
\hline
Intervention/treatment and control/placebo &
``Experimental arm: 2 mg of the study drug subcutaneously once weekly; comparator: matching placebo.'' [\texttt{PREREG\_0009}] &
``Participants received 2 mg of the study drug or matching placebo subcutaneously once weekly, in addition to standard care.'' [\texttt{PAPER\_0015}] &
\textbf{No deviation.} Drug, dose, route, schedule, and comparator all match. \\
\hline
Ethical approval -- number &
The registry record does not contain an ethics approval number. &
``The study was approved under protocol number 2020-417.'' [\texttt{PAPER\_0003}] &
\textbf{Insufficient evidence.} The registry records no approval number, so the two cannot be compared. \\
\hline
Ethical approval -- committee &
``Oversight: Institutional Review Board of the coordinating centre.'' [\texttt{PREREG\_0002}] &
``The trial was approved by the Institutional Review Board of the coordinating centre.'' [\texttt{PAPER\_0004}] &
\textbf{No deviation.} Both identify the same ethics committee. \\
\hline
Ethical approval -- date &
No ethics approval date is recorded in the registry. &
``Ethical approval was obtained prior to enrolment.'' [\texttt{PAPER\_0005}] &
\textbf{Insufficient evidence.} Neither source states a specific approval date. \\
\hline
Sample size &
``Estimated enrolment: 300 participants, randomised 1:1.'' [\texttt{PREREG\_0007}] &
``The trial was terminated early due to slow recruitment; 212 of a planned 300 participants underwent randomisation.'' [\texttt{PAPER\_0014}] &
\textbf{Deviation.} Enrolment fell short of the registered target and the trial was stopped early. \\
\hline
Date recruitment started &
``Study start date: March 2021.'' [\texttt{PREREG\_0008}] &
``Recruitment began in March 2021 and ended in November 2022.'' [\texttt{PAPER\_0013}] &
\textbf{No deviation.} The reported recruitment start matches the registered start date. \\
\hline
Outcomes -- primary &
``Primary outcome measure: change in HbA1c from baseline to Week 24.'' [\texttt{PREREG\_0011}] &
``The primary endpoint was the proportion of participants achieving an HbA1c below 7.0\% at Week 24.'' [\texttt{PAPER\_0019}] &
\textbf{Deviation.} The registered continuous primary outcome was replaced by a dichotomous responder endpoint and was not disclosed as a change. \\
\hline
Outcomes -- secondary &
``Secondary outcome measure: change in body weight from baseline to Week 24.'' [\texttt{PREREG\_0021}] &
No excerpt addressing body weight was retrieved from the manuscript text. &
\textbf{Insufficient evidence.} The registered secondary outcome could not be located in the parsed manuscript. \\
\hline
Method of randomisation and allocation &
``Allocation: randomised; computer-generated sequence; 1:1; central allocation.'' [\texttt{PREREG\_0010}] &
``Participants were randomly assigned 1:1 using a computer-generated sequence with central allocation concealment.'' [\texttt{PAPER\_0017}] &
\textbf{No deviation.} The described randomisation and allocation match the registration. \\
\hline
\end{tabular}
\end{table}

Two things can be immediately noted in this report. First, the presence of each row immediately provides structure for the user in how to think about comparing the registration and manuscript. Second, the distinct rows fall into categories which demand one of three courses of action.

RegCheck states that the paper is consistent with the registration on five dimensions (\textbf{Eligibility - inclusion criteria}, the \textbf{Intervention/treatment
and control/placebo}, \textbf{Ethical approval - committee}, \textbf{Date recruitment
started}, and \textbf{Method of randomisation and allocation}). Each of these is a row the reviewer can confirm at a glance, following the evidence identifiers to the matching text in the registry record and the manuscript. They additionally can scrutinise these consistency judgements further by easily viewing the surrounding context of these quotes around the extracted text to verify that RegCheck did not miss anything important. 

Three dimensions are flagged as ``deviations'', and they differ instructively in their visibility: 
\begin{enumerate}
    \item The \textbf{Outcomes – primary} row exhibits a clear deviation, with the outcome registered as continuous (change in HbA1c) but reported in the paper as a dichotomous responder endpoint. Such outcome switching is at odds with CONSORT reporting standards, can alter the apparent strength of the findings, and is not disclosed as a change. At the same time, this subtle difference can be easy to miss in manual review: RegCheck facilitates this comparison by flagging it trivially. 
    \item The \textbf{Eligibility – exclusion criteria} row is an even subtler case, exhibiting a renal-function threshold changed from 30 to 45 mL/min -- yet again exactly the kind of small numeric change that is almost impossible to catch by eye but that can be trivially spotted and verified by RegCheck. 
    \item The \textbf{Sample size} row catches a case of an early termination, as well as under-enrolment relative to the registered target. In each of these cases, RegCheck flags these issues because they are literal deviations, but RegCheck should not be used to outsource expert decision-making. Indeed, each of these departures may well be defensible with further expert scrutiny. However, they depart from what was registered, and are therefore flagged so that an expert can review them accordingly.
\end{enumerate}

The remaining three rows return ``insufficient evidence'' for different reasons. For \textbf{Ethical approval – number} and \textbf{Ethical approval – date}, the registry simply does not carry the information, so there is nothing on the registration side to compare against. RegCheck simply reports this absence of information directly and then moves on. For \textbf{Outcomes - secondary}, the registration may contain this information, but the relevant text could not be located in the parsed manuscript. Rather than assert that the outcome was dropped, RegCheck directs the reviewer to confirm whether it is reported elsewhere (e.g., in a supplement not provided to RegCheck).

In each case, RegCheck has distilled a slow, full-document reading task to a short, structured set of extractions and judgements that can be reviewed, verified, or rejected in mere minutes. Indeed, this is precisely the use case of RegCheck: to serve an orienting function to researchers and make a registration-paper comparison manageable enough to actually perform in an already cumbersome review workflow. The same applies to authors themselves, who can run RegCheck on their own manuscripts before submission to catch and disclose undetected deviations proactively. Importantly, RegCheck leaves the substantive judgements (e.g., whether or not observed deviations are threatening to the integrity of the research claims) where they belong: with the human expert.

\section{Use Cases}
The most obvious use case for RegCheck is for editors and reviewers who are assessing a manuscript where one or more studies were registered (where they have permission to share and upload those materials; see Privacy and Confidentiality). Similarly, research synthesists might use it when assessing the trustworthiness of studies included in a systematic review or meta-analysis, given the growing emphasis such checks from organisations such as Cochrane \citep{wilkinson_inspect-sr_2025}. Indeed, INSPECT-SR \citep[the Cochrane-endorsed checklist for conducting trustworthiness checks in systematic reviews;][]{wilkinson_inspect-sr_2025} focuses a portion of its checks on registration-trial consistency, and RegCheck will be included as an endorsed tool in the next iteration of INSPECT-SR. Of course, RegCheck is also useful for authors who want to check their own manuscripts for any deviations they should report \citep[e.g., RegCheck could help to spot registration deviations that might be missing in a registration deviation table, like the one suggested by][]{willroth_best_2024}. Regardless of whether the user is using RegCheck to assess their own paper or someone else's, the process is the same.

\section{Design Principles and Philosophy}
\begin{itemize}
\item \textbf{Pragmatism.} RegCheck is a software tool which provides a standardised framework for checking registrations against papers, with the goal of increasing the frequency of registration-paper comparisons (which is not done regularly). 
\item \textbf{Human-in-the-loop.} RegCheck facilitates human experts; it does not replace them. The system highlights candidate discrepancies, and also provides indicative LLM-based judgements, but it is not intended to substitute human expertise.
\item \textbf{Non-prescriptive.} It is not the role of RegCheck to make determinations about which dimensions matter when comparing a registration to a paper: this can vary by user, use case, and field/discipline norms. There is no broad cross-discipline consensus, that we know of, on the selection of dimensions which do or do not matter for study registration. RegCheck therefore provides discipline-specific default dimensions based on frequently examined features (e.g., sample size, primary outcomes), spanning a range of fields, while also providing free-text fields for users to add or redefine dimensions of their own.
\item \textbf{Discipline-agnostic and discipline-aware.} Although it has origins in psychological research, RegCheck is an abstract workflow aimed at being useful for researchers across a range of scientific disciplines, and supports distinct comparison flows for different registration types (for example, preregistrations, clinical-trial registrations, and preclinical/animal-study registrations). At the same time, we recognise that RegCheck's accuracy and usefulness may vary across fields and disciplines; for this reason, we are investigating variation in its accuracy in a range of different domains (see the `Future Directions' section below).
\item \textbf{Modular separation of roles.} Because RegCheck is built as a specific instance of Parity's research artefact comparison, it explicitly separates document ingestion, human definition of dimensions, similarity-based evidence extraction, and model-based adjudication. This makes the system easier to audit, evaluate, and extend, and it clarifies that errors in one part of the workflow need not be attributed wholesale to ``the LLM''. RegCheck’s modularisation also enables it to keep pace with the rapid release of new models and tools in this fast-evolving space. By structuring the workflow through the IDEA framework, individual components can be readily updated or replaced as improved models become available.
\end{itemize}

\section{Privacy and Confidentiality}
To create the persistent, shareable report page, RegCheck stores the artefacts used to create the report: namely, the text excerpts and quotes extracted for each dimension, the rendered document pages that are shown in the Evidence view, and a copy of the source documents (i.e., registration/paper) used to render that view. For reports generated anonymously (i.e., by a user who is not logged in), these artefacts are retained only for a limited period (approximately 7 days) and are then deleted automatically. For signed-in users, RegCheck stores generated reports indefinitely, but also offers users the ability to delete their generated reports at any time. 

RegCheck provides users with the option to decide which generative LLM is used in the creation of the RegCheck report. Some of these models are open-weights models which are hosted on third-party inference API providers (e.g., the Groq service). Others are closed-source models which are hosted by those commercially invested in that model (e.g., OpenAI serves its own models). LLM providers typically state in their documentation that text served to its models via its API service is not retained nor used for training. However, we of course have no control over the enforcement of, and adherence to, these privacy policies, nor do we have the capacity to know whether such policies may change in the future. We therefore provide the open-weights models as alternative options, which are hosted by third-party providers who are not commercially invested in the development of the models they host. RegCheck can also be run fully locally, either via its CLI, or by forking the open-source repository and running the service locally. 

Usage of RegCheck should be constrained by the same standards of confidentiality and privacy around academic materials that apply in any other professional context. In other words: just as one should not share confidential materials with unauthorised colleagues, one should not upload such materials to RegCheck. 

\section{Accounts, Sharing, and Programmatic Access}
RegCheck can be used entirely anonymously, but users may create an account (using a supported identity provider, such as Google or ORCID) to use a small set of collaboration features. Signed-in users can name and retain the reports they generate, browse them from a personal dashboard, and control who can view each report. Report visibility is two-tiered: a report can be made public, so that anyone with its RegCheck ID link can open it, or kept private, in which case access is restricted to an explicit allow-list of collaborators (with a RegCheck account) identified by verified email address or ORCID iD. This makes it possible, for example, for an editor to share a confidential comparison only with the assigned reviewers. For programmatic and higher-volume use, RegCheck also exposes an HTTP API authenticated with per-user API keys, mirroring the comparison capabilities of the web app so that comparisons can be scripted or integrated into editorial and meta-research pipelines.

\section{Running RegCheck Online and Locally}
RegCheck is freely available to use. Its point-and-click web application is available at \href{https://regcheck.app}{https://regcheck.app}, and the open-source code, which can be run locally at one's own cost, is available on \href{https://github.com/JamieCummins/regcheck}{GitHub}. The cost of running a comparison varies with the size of the registration and paper and with the selected model; for users of the public web application, these costs are currently covered by our research group, and use is free. 

Beyond its web app, RegCheck comes with a headless command-line interface (CLI) for running comparisons entirely locally without any hosted infrastructure. The CLI covers the same comparison flows as the web app, takes comparison dimensions either from the built-in discipline presets or from a user-supplied file, and can be used with either hosted or fully local model providers. This makes it a natural path for privacy-sensitive use: paired with a local embedding endpoint and a locally hosted language model, an entire comparison can be run with no document or text leaving the user's environment. The CLI can write the same interactive report as a single self-contained offline HTML file, so that a report can be archived or shared as a file rather than as a link. The CLI also offers an additional feature, \textit{consensus mode}, that re-runs the adjudication step several times per dimension. The final judgement reported is the verdict representing the plurality of these calls. This offers protection against the stochasticity of LLM outputs, which can be particularly problematic for smaller models that will likely be used in local workflows.

\section{Accuracy and Validation}
RegCheck's provided judgement of deviations, and deviation information, can additionally be useful for spotting deviations that may have been overlooked. However, RegCheck is not intended to replace human judgement. Users must apply their own discretion when evaluating registration-paper consistency, and they should be aware that RegCheck can make both false positive errors (cases where RegCheck flags that there is a deviation when there is not a deviation; also keep in mind that very minor deviations may be flagged) and false negative errors (cases where RegCheck flags that there is no deviation when there is a deviation). 

We do not yet provide guarantees about the fidelity of RegCheck's deviation judgements, and we strongly advise users to verify all output for accuracy while systematic evaluation is ongoing. With this said, our impressions, based on extensive testing among our research team and close colleagues, are that RegCheck is generally very accurate in its extractions and adept at catching genuine deviations where they exist. This has frequently included deviations that members of our research team missed, such as subtle differences in registered vs.\ reported numeric inclusion-criteria thresholds, small changes in planned analyses, and instances of outcome-switching in clinical trials. However, anecdotal testing is not a substitute for formal evaluation. RegCheck research therefore focuses on quantifying performance against human-coded data for comparisons of registrations and papers in psychology, (pre)clinical medicine, and economics.

Evaluating tools like RegCheck comes with challenges. Importantly, there is little consensus on how one can assess such performance in humans, let alone automated tools. This is largely due to two issues. First, there is no agreed-upon framework on how to evaluate registration-paper consistency: different researchers disagree about what should be specified in a registration, what dimensions along which a registration should be compared to a paper, and what ultimately counts as a deviation or not. Second, some aspects of comparison are inherently ambiguous: whereas it can be relatively straightforward to get agreement on a prespecified sample size, other features of comparison (e.g., prespecified hypotheses or sufficiency of prespecified outcome descriptions) can elicit more variation in the evaluations of humans.

Put simply: in many cases, establishing a ground truth among human comparators is incredibly difficult, and we lack systematic frameworks for making such comparisons in the first place. In evaluating RegCheck, we acknowledge this fact, and use it to inform our evaluation approach. Specifically, rather than trying to evaluate RegCheck as a software aimed at approximating a known ground-truth value, we instead treat and evaluate it as an accompanying reviewer. In practice, this means one of our core evaluation metrics of interest is the degree of agreement with other human comparators, and the establishment of relative non-inferiority to observed human-human agreement. In other words: for us to consider RegCheck sufficiently useful, it should, in general, show agreement with human raters to at least the same extent as humans show agreement with one another. More concretely, our goals in evaluating RegCheck are in terms of interrater agreement on (i) extracted text from registrations/papers (e.g., planned research questions, planned statistical tests), and (ii) judgements of consistency between the registration and paper. To do this for a given research area, we specify and define a set of dimensions along which registrations and papers should be compared, solicit multiple human comparators to provide these comparisons, and examine agreement among these human evaluations and those of RegCheck. In cases of disagreement, we will also examine whether RegCheck is hallucinating or making mistakes, or capturing legitimate details which humans tend to miss.

Although our primary investigations focus on the aspect of RegCheck-human consistency, we additionally plan to examine RegCheck's accuracy through artificially generating registration and paper texts using a generative language model which contain specific injected inconsistencies. Such an approach is more amenable to conventional means of evaluation, given that we are in control of the ground-truth consistency in these cases. This approach also provides us with opportunities to conduct evaluation at greater scales: whereas soliciting human comparators to review registration-paper consistency takes a substantial deal of time, evaluating RegCheck's performance on artificially generated examples can be done much more quickly and efficiently. Of course, the drawback to this is that the artificial texts may be relatively generic, and written in ``LLM-speak'' which will likely not be representative of the diversity of writing styles present in the true academic literature. Consequently, these diverging approaches (interrater agreement with human comparators, artificially generated texts with known (in)consistencies) provide us with multiple perspectives on the accuracy of RegCheck: both its conventional accuracy in an artificial context, as well as its accuracy in more realistic settings.

Of course, in principle RegCheck could still serve a pragmatic function even if it exhibits lower interrater agreement than that observed for human-human agreement. After all, the baseline rate of conducting registration-paper comparisons in typical human reviewing is low, and this is largely due to the fact that doing such comparisons is time-consuming and laborious. Checking the accuracy of an existing RegCheck report, by contrast, is much less time-consuming. In this sense, RegCheck may serve an orienting function to authors and reviewers. Even if its extractions are at times inaccurate, it may functionally increase the likelihood of these checks being done in the first place, where currently the baseline rate of this behaviour is very low. However, this is obviously not optimal: our goal with RegCheck is to provide refined software whose text extractions and consistency judgements reach the same standard of agreement with humans as those of another human.

\section{Future Directions}
RegCheck is best seen as infrastructure to support robust and trustworthy science. By making registration-paper comparison more efficient, more consistent, and more transparent, we hope to make such comparisons routine in authors' own workflows as well as in editorial workflows, peer review, and post-publication auditing. Its modularity allows adaptation across disciplines and integration into larger ecosystems for research integrity.
The research arm of RegCheck's development focuses on concretely addressing:
\begin{enumerate}
\item The accuracy of RegCheck's text extraction and deviation judgments, and whether this accuracy varies across domains, disciplines, and model hyperparameters;
\item Whether providing RegCheck to authors can improve the quality of registration-paper comparisons upon submission of papers to journals;
\item Whether RegCheck improves the efficiency and frequency of registration-paper comparisons.
\end{enumerate}
We welcome engagement and input from researchers, journals, and funders interested in using RegCheck in diverse research areas. We are working to develop extensions of RegCheck, including the capacity to (i) assess multiple papers at once for use in meta-science or meta-analysis research, and (ii) assess consistencies and deviations between study analysis code and descriptions of planned analyses.

\section{Conclusion}
Study registration promises to make the difference between planned and post hoc decisions traceable, and to help readers calibrate their confidence in a result. That promise is only realised when registrations are actually checked against the papers they accompany. As empirical investigations demonstrate, this checking is rarely done. One reason for this is that manual checking is slow, repetitive, and easy to defer. One might argue that researchers can now simply upload these documents to a large language model and prompt it to compare them: however, this is necessarily unstandardised, both in terms of the specific set-up provided to the LLM and the criteria along which the LLM may compare the two materials. 

RegCheck provides a systematic and structured approach to automated comparisons of registrations and papers as a direct response to this gap. By using the conceptual scaffolding of the IDEA framework, it decomposes the task into four modular steps, enabling each step to be independently evaluated, and improved. This also serves to delineate clearly where generative LLMs are and are not required: their primary purpose is in the adjudication step, where synthesised judgement is required.

The ultimate goal of RegCheck is to support researchers in engaging in an otherwise rare scientific activity: comparing registrations to papers. Used in this spirit, we believe RegCheck (and the Parity architecture behind it) can help make this consistency checking a more routine part of the scientific writing and review process in the social and medical sciences.

\section*{Declarations}
\subsection*{Ethics approval and consent to participate}
Not applicable.
\subsection*{Consent for publication}
Not applicable.
\subsection*{Availability of data and materials}
RegCheck is open source; code is available at \url{https://github.com/JamieCummins/regcheck} under the GNU AGPL-3.0 licence. 
\subsection*{Competing interests}
Beyond affiliation with the (not-for-profit, free-to-use) RegCheck software, the authors declare no competing interests. 
\subsection*{Funding}
JC, BC, and ME are supported by a joint grant in the META-REP Priority Program of the German Research Foundation (\#546323839) and the Swiss National Science Foundation (\#100014E\_225145).
\subsection*{Authors' contributions}
CRediT: Conceptualization: JC, BC, IH, ME; Funding acquisition: ME; Methodology: JC, BC, IH, ME; Project administration: JC, BC; Resources: JC, ME; Software: JC; Supervision: ME; Writing – original draft: JC; Writing – review \& editing: JC, BC, IH, ME.
\subsection*{Acknowledgements}
We thank (in alphabetical order) Alison Avenell, Anouk Bouma, Saloni Dattani, Matilda Fogato, Lukas Jung, Daniel Lakens, Alexander Mainetti, Sabrina Norwood, Lars Schilling, Olmo van den Akker, Colby Vorland, and Alexander Wuttke for providing valuable feedback on RegCheck's user experience and output during the course of its development.

\bibliographystyle{apacite}
\bibliography{references}

\end{document}